\documentclass[letterpaper]{article} % DO NOT CHANGE THIS
\usepackage{aaai24}  % DO NOT CHANGE THIS
\usepackage{times}  % DO NOT CHANGE THIS
\usepackage{helvet}  % DO NOT CHANGE THIS
\usepackage{courier}  % DO NOT CHANGE THIS
\usepackage[hyphens]{url}  % DO NOT CHANGE THIS
\usepackage{graphicx} % DO NOT CHANGE THIS
\usepackage{natbib}  % DO NOT CHANGE THIS AND DO NOT ADD ANY OPTIONS TO IT
\usepackage{caption} % DO NOT CHANGE THIS AND DO NOT ADD ANY OPTIONS TO IT
\usepackage{booktabs}       % professional-quality tables
\usepackage{amsfonts}       % blackboard math symbols
\usepackage{nicefrac}       % compact symbols for 1/2, etc.
\usepackage{microtype}      % microtypography
\usepackage{xcolor}         % colors
\usepackage{algorithm}
\usepackage{algorithmic}
\usepackage{tabulary}
\usepackage{textcomp}
\usepackage{graphicx}
\usepackage{subfigure}
\usepackage{multirow}

\usepackage{newfloat}
\usepackage{listings}
\DeclareCaptionStyle{ruled}{labelfont=normalfont,labelsep=colon,strut=off} % DO NOT CHANGE THIS
\floatstyle{ruled}
\newfloat{listing}{tb}{lst}{}
\floatname{listing}{Listing}
\title{Decoupled  Training:\\ Return of Frustratingly Easy Multi-Domain Learning}
\author {  
Ximei Wang\textsuperscript{\rm 1}\thanks{Corresponding author.},
Junwei Pan\textsuperscript{\rm 1},
Xingzhuo Guo\textsuperscript{\rm 2},
Dapeng Liu\textsuperscript{\rm 1}, 
Jie Jiang\textsuperscript{\rm 1}
}
\affiliations {
\textsuperscript{\rm 1}Tencent Inc, China \\
\textsuperscript{\rm 2}School of Software, Tsinghua University, China\\
\{messixmwang,jonaspan,rocliu,zeus\}@tencent.com, gxz23@mails.tsinghua.edu.cn
}

\usepackage{bibentry}
\begin{document}

\maketitle

\begin{abstract}
	Multi-domain learning (MDL) aims to train a model with minimal average risk across multiple overlapping but non-identical domains. To tackle the challenges of dataset bias and domain domination, numerous MDL approaches have been proposed from the perspectives of \textit{seeking commonalities} by aligning distributions to reduce domain gap or \textit{reserving differences} by implementing domain-specific towers, gates, and even experts. MDL models are becoming more and more complex with \emph{sophisticated} network architectures or loss functions, introducing extra parameters and enlarging computation costs. In this paper, we propose a \emph{frustratingly easy} and hyperparameter-free multi-domain learning method named Decoupled Training (D-Train). D-Train is a tri-phase \emph{general-to-specific} training strategy that first pre-trains on all domains to warm up a root model, then post-trains on each domain by splitting into multi-heads, and finally fine-tunes the heads by fixing the backbone, enabling decouple training to achieve domain independence. Despite its extraordinary simplicity and efficiency, D-Train performs remarkably well in extensive evaluations of various datasets from standard benchmarks to applications of satellite imagery and recommender systems.
\end{abstract}

\section{Introduction}
\label{intro}
The success of deep learning models across a wide range of fields often relies on the fundamental assumption that the data points are \textit{independent and identically distributed} (i.i.d.). However, in real-world scenarios, training and test data are usually collected from different regions or platforms, consisting of multiple overlapping but non-identical domains. For example, a popular satellite dataset named FMoW ~\cite{christie2018functional}, which predicts the functional purpose of buildings and land use on this planet, contains large-scale satellite images from various regions with different appearances and styles. Thus, jointly training a single model obscures domain distinctions, while separately training multiple models by domains reduces training data in each model~\cite{when_do_domain_matters}. This dilemma motivated the research on \textit{multi-domain learning}~\cite{when_do_domain_matters, liu2017adversarial, ma2018modeling, MulANN, tang2020progressive}.

\begin{figure*}[tp]
	\centering
	\subfigure[Review Examples of MDL]{
		\label{review_example}
		\includegraphics[width=0.21\textwidth]{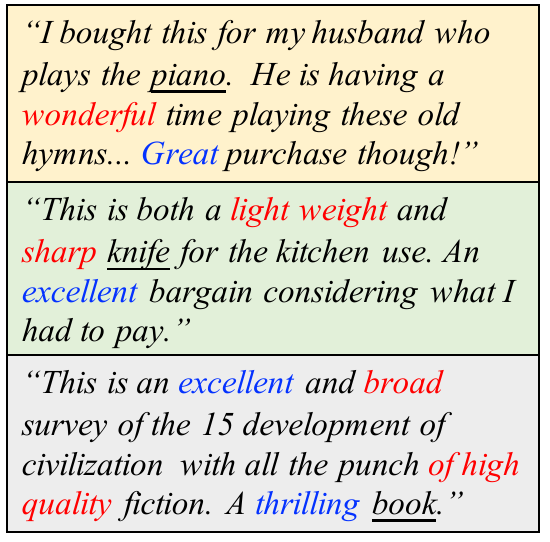}
	}
	\subfigure[Visual Examples  of MDL]{
		\label{office_home_samples}
		\includegraphics[width=0.21\textwidth]{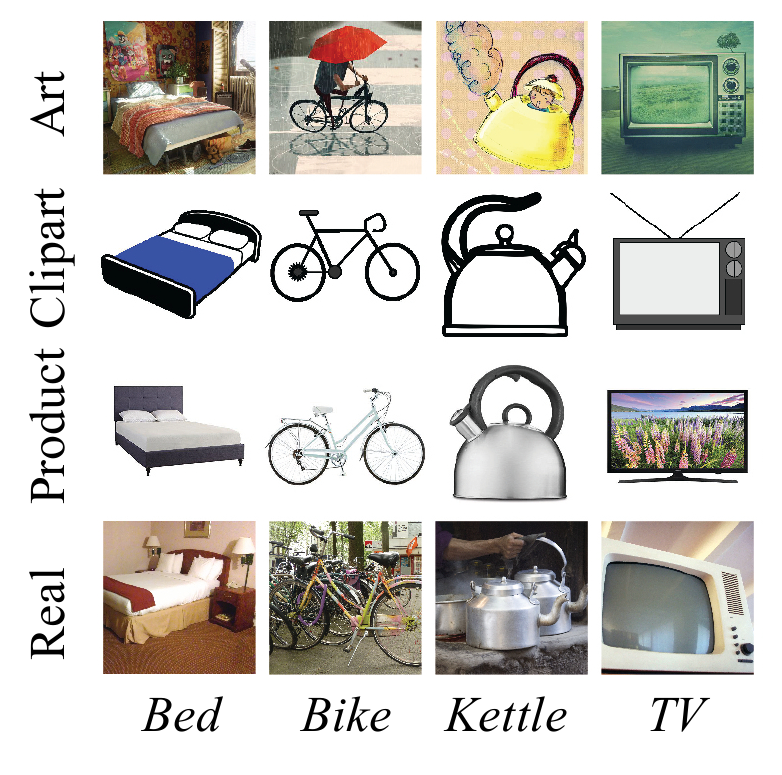}
	}
	\subfigure[\#Samples in Amazon]{
		\label{TwistedShift}
		\includegraphics[width=0.25\textwidth]{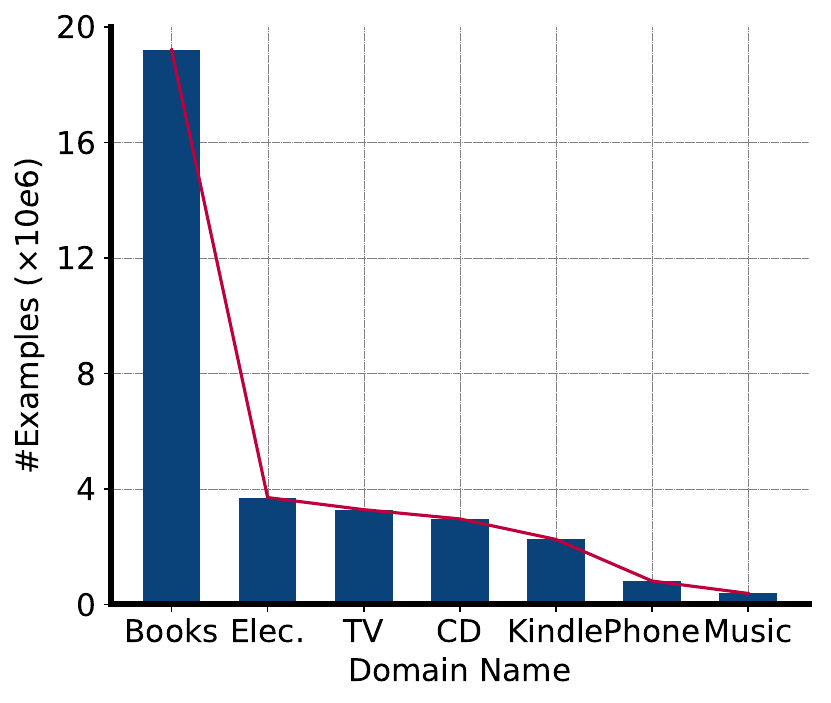}
	}
	\subfigure[\#Samples in FMoW]{
		\label{FMoW_examples}
		\includegraphics[width=0.25\textwidth]{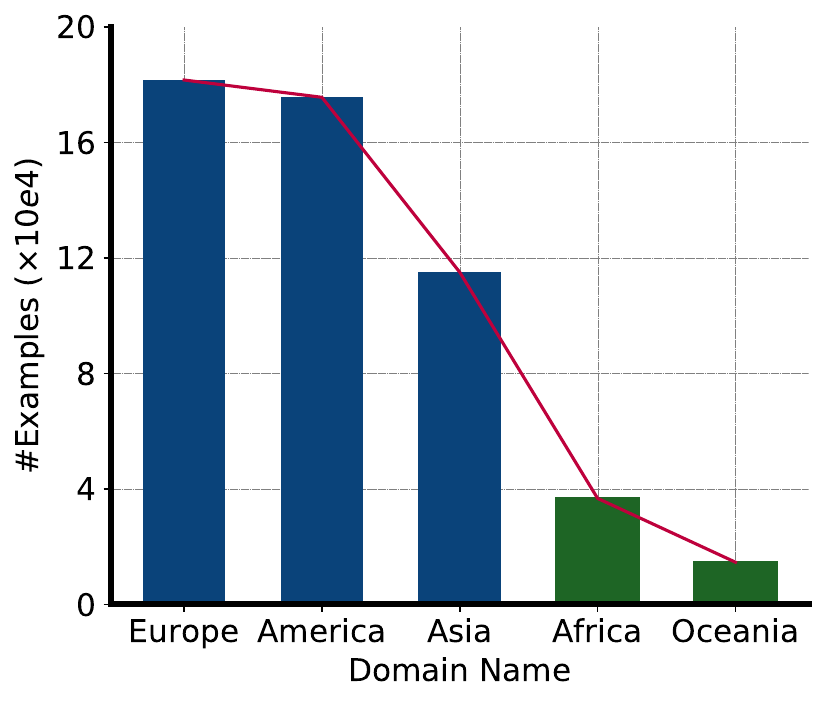}
	}
	\caption{{(a)}-{(b)}: Review examples from a recommender system benchmark named Amazon with various styles and keywords (shown by different colors), as well as visual examples from a computer vision dataset named Office-Home with various appearances and backgrounds, reveal the main challenge of {dataset bias}.
		{(c)}-{(d)}: The distribution of sample number across domains is naturally {imbalanced} or even long-tailed, indicating another major challenge of {domain domination}.}
	\label{fig:distribution}
\end{figure*}

\begin{figure*}[tp]
	\centering
	\subfigure[Seeking Commonalities]{
		\label{seeking_common}
		\includegraphics[width=0.504\textwidth]{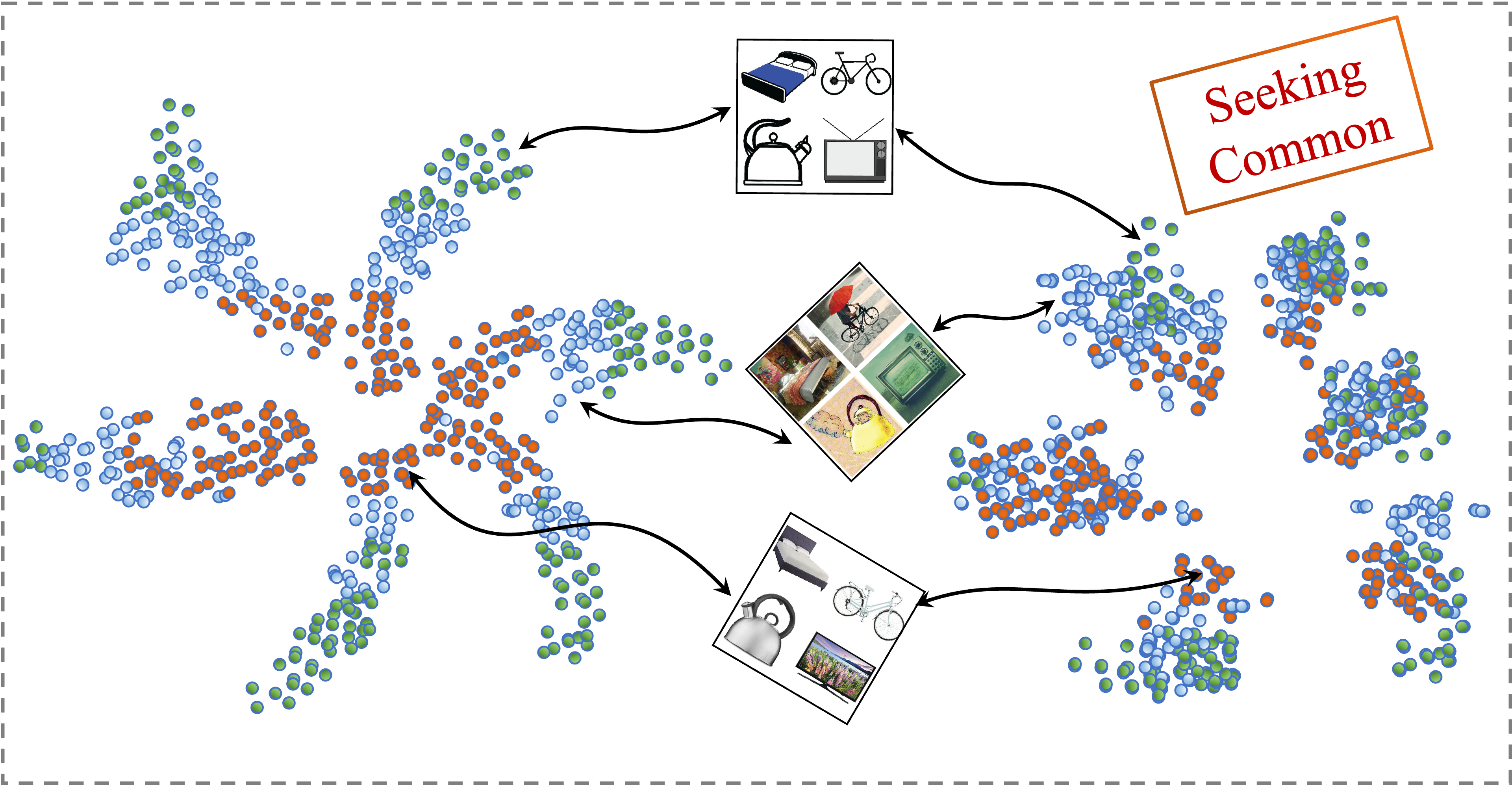}
	}
	\subfigure[Reserving Differences]{
		\label{reserving_differences}
		\includegraphics[width=0.36\textwidth]{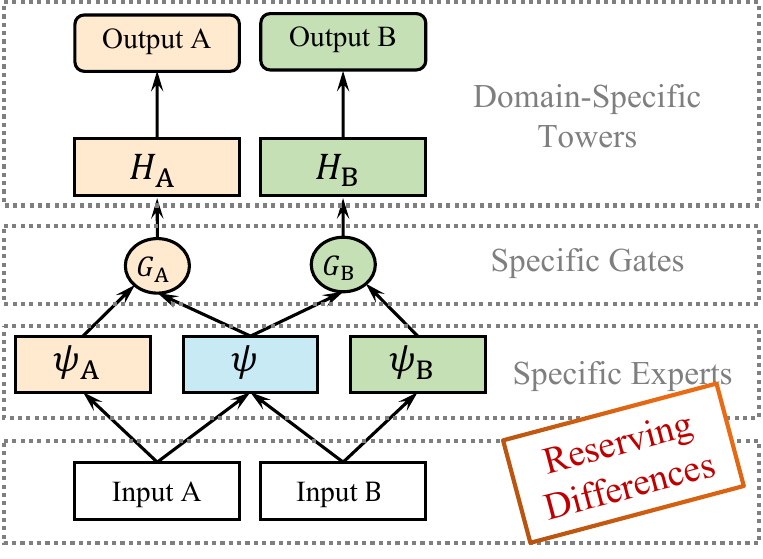}
	}
	\caption{{(a)}: Seeking Commonalities by aligning distributions across domains to reduce domain gap. {(b)}: Reserving Differences by implementing domain-specific towers, gates, and even experts.}
	\label{fig:compare}
\end{figure*}

To figure out the challenges of multi-domain learning, we first delved into these two standard benchmark datasets. By analyzing the examples of these datasets, it is obvious that \textit{dataset bias} across domains is one of the biggest obstacles to multi-domain learning. As shown in Figure~\ref{review_example}, reviews from different domains have different keywords and styles. Further, Figure~\ref{office_home_samples} reveals that images of Office-Home are from four significantly different domains: \textit{Art}, \textit{Clipart}, \textit{Product} and \textit{Real-World}, with various appearances and backgrounds. To tackle the \textit{dataset bias} problem, numerous approaches have been proposed and they can be briefly grouped into two categories: 1) \textbf{Seeking Commonalities}. Some classical solutions~\cite{ liu2017adversarial, MulANN} adopting the insightful idea of domain adversarial training have been proposed to extract domain-invariant representations across multiple domains. 2) \textbf{Reserving Differences}. These approaches ~\cite{ma2018modeling, tang2020progressive} adopt multi-branch network architectures with domain-specific towers, gates, and even experts,  implementing domain-specific parameters to avoid domain conflict caused by dataset bias across domains. As shown in Figure~\ref{fig:compare}, these methods are becoming more and more complex with \textit{sophisticated} network architectures or loss functions.

Further, different from multi-task learning which focuses on tackling different tasks within a single domain, multi-domain learning shares the same label space across multiple domains with different marginal distributions. We thus calculated the distribution of sample amount across domains on several benchmark datasets and the results are shown in Figure~\ref{fig:distribution}. In practice, the distributions of sample amount across domains are usually twisted, imbalanced, or even long-tailed, causing another main challenge of multi-domain learning: \textit{domain domination}. In this case, some tailed domains may have much fewer examples and it would be difficult to train satisfied models on them. Without proper network designs or loss functions, the model will be easily dominated by the head domains with many more examples and shift away from the tailed ones with few examples.

Realizing the main challenges of \textit{dataset bias} and \textit{domain domination} in multi-domain learning, we aim to propose a general, effective, and efficient method to tackle these obstacles all at once.  We first rethought the development of multi-domain learning and found that the approaches in this field are becoming more and more sophisticated, consisting of multifarious network architectures or complex loss functions with many trade-off hyperparameters. By assigning different parameters across domains, these designs may be beneficial in some cases but they will include more parameters and enlarge computation cost,  as well as introduce much more hyperparameters. For example, the latest state-of-the-art method named PLE has to tune the numbers of domain-specific experts and domain-agnostic experts in each layer and design the network structures of each expert, each tower, and each gate network.

Motivated by the famous quote of Albert Einstein, 	
``\textit{everything should be made as simple as possible, but no simpler}'', we proposed a frustratingly easy multi-domain learning method named Decoupled Training (D-Train). D-Train is a training strategy based on the original frustratingly easy but most general \emph{shared-bottom} architecture. D-Train is a tri-phase \emph{general-to-specific} training strategy that first pre-trains on all domains to warm up a root model, then post-trains on each domain by splitting into multi-heads, and finally fine-tunes the heads by fixing the backbone, enabling decouple training to achieve domain independence. Despite its extraordinary simplicity and efficiency, D-Train performs remarkably well in extensive evaluations of various datasets from standard benchmarks to applications of satellite imagery and recommender systems. In summary, this paper has the following contributions:

\begin{itemize}
	\item We explicitly uncover the main challenges of dataset bias and domain domination in MDL, especially the latter since it is usually ignored in most existing works.
	\item We propose a frustratingly easy and hyperparameter-free MDL method named Decoupled Training by applying a tri-phase \emph{general-to-specific} training strategy.
	\item We conduct extensive experiments from standard benchmarks to real-world applications and verify that D-Train performs remarkably well. 
\end{itemize}

\section{Related Work}

\subsection{Seeking Commonalities}

To tackle the \textit{dataset bias} problem in multi-domain learning, various approaches~\cite{ liu2017adversarial, MulANN} have been proposed from the perspective of domain alignment, by adopting the insightful idea of domain adversarial training to extract domain-invariant representations across domains. There are mainly two categories of domain adaptation formulas: \textit{covariate shift}~\cite{dataset_shift_in_ML09,pan2010survey,Long15DAN,ganin15RevGrad} and \textit{label shift}~\cite{Lipton18ICML,Azizzadenesheli19ICLR,alexandari2020maximum}, while we focus on the former in this paper since it is more relevant with the topic of MDL. Recent deep domain adaptation methods tackle domain shifts from the perspectives of either moment matching or adversarial training, in which the former aligns feature distributions by minimizing the distribution discrepancy across domains~\cite{Long15DAN, Tzeng14DDC, Long17JAN}.
Further, domain adversarial neural network (DANN) \cite{ganin2016domain} becomes the mainstream method in domain adaptation. It introduces a domain discriminator to distinguish the source features from the target ones, while the feature extractor is designed to confuse the domain discriminator. In this way, the domain discriminator and feature extractor are competing in a two-player minimax game. Its natural extension to MDL is DANN-MDL. Later, CDAN \cite{Long18CDAN} further tailors the discriminative information conveyed in the classifier predictions into the input of the domain discriminator, whose natural extension to MDL is CDAN-MDL.
Following the main idea of the minimax game, several variants of adversarial training methods~\cite{Pei18MADA, Tzeng17ADDA, Saito17MCD, ICML2019MDD} were proposed. MulANN~\cite{MulANN} tailors the insight of domain adversarial training into the MDL problem by introducing a domain discriminator into the shared model with a single head. In contrast, ASP-MTL~\cite{liu2017adversarial} includes a shared-private model and a domain discriminator.

\subsection{Reserving Differences}

Another series of methods~\cite{ma2018modeling, tang2020progressive, STAR21CIKM}  for multi-domain learning adopt multi-branch network architectures and develop domain-specific parameters~\cite{Dredze10ConfidenceWeightedCombination} to avoid domain conflict caused by dataset bias across domains.  Among them, Shared Bottom (SB)~\cite{ruder2017overview} is the frustratingly easy but effective one. Further, MoE~\cite{jacobs1991adaptive} and its extension of MMoE~\cite{ma2018modeling} adopt the insightful idea of the mixture of experts to learn different mixture patterns of experts assembling, respectively. PLE~\cite{tang2020progressive} explicitly separates domain-shared and domain-specific experts to alleviate harmful parameter interference across domains. Note that, PLE further applies progressive separation routing with several deeper layers but we only adopt one layer for a fair comparison with other baselines. Other MDL methods focus on maintaining shared and domain-specific parameters by confidence-weighted combination~\cite{Dredze10ConfidenceWeightedCombination}, domain-guided dropout~\cite{Xiao16DomainGuidedDropout}, or prior knowledge about domain semantic relationships~\cite{ICLR15Unified}. Meanwhile, various task-specific MDL approaches have been proposed for computer vision~\cite{rebuffi2018efficient, mancini2020boosting, nam2016learning, rebuffi2017learning, li2019efficient, fourure2017multi}, natural language processing~\cite{wu2020dual, williams2013multi, pham2021revisiting} and recommender system~\cite{hao2021adversarial,chen2020scenario,du2019sequential,gu2021self, li2021debiasing, li2021dual, salah2021towards}. The comparison between these MDL methods is summarized in  Figure~\ref{fig:compare}.

\section{Approach}

This paper aims to propose a simple and effective method for MDL. Given data points from multiple domains $\{\mathcal{D}_{1}, \mathcal{D}_{2},..., \mathcal{D}_{T}\}$, $T$ is the domain number. Denote $\mathcal{D}_t = \{(\mathbf{x}_i,\mathbf{y}_i)\}_{i=1}^{n_t}$ the data from domain $t$,  where $\mathbf{x}_i$ is an example, $\mathbf{y}_i$ is the associated label and $n_t$ is the sample number of domain $t$. Denote the shared backbone $\psi$ and the domain-specific heads $\{{h}_{1}, {h}_{2},..., {h}_{T}\}$ respectively. The goal of MDL is to improve the performance of each domain $\mathcal{D}_t$.

As mentioned above, the proposed Decoupled Training (D-Train) is a frustratingly easy multi-domain learning method based on the original \emph{shared-bottom} architecture. D-Train takes a \emph{general-to-specific} training strategy. It first pre-trains on all domains to warm up a root model. Then, it post-trains on each domain by splitting into multi-heads. Finally, it fine-tunes the heads by fixing the backbone, enabling decouple training to achieve domain independence.

\begin{figure*}[thbp]
	\centering
	\subfigure[Training details of each phase]{
		\label{phase}
		\includegraphics[width=0.45\textwidth]{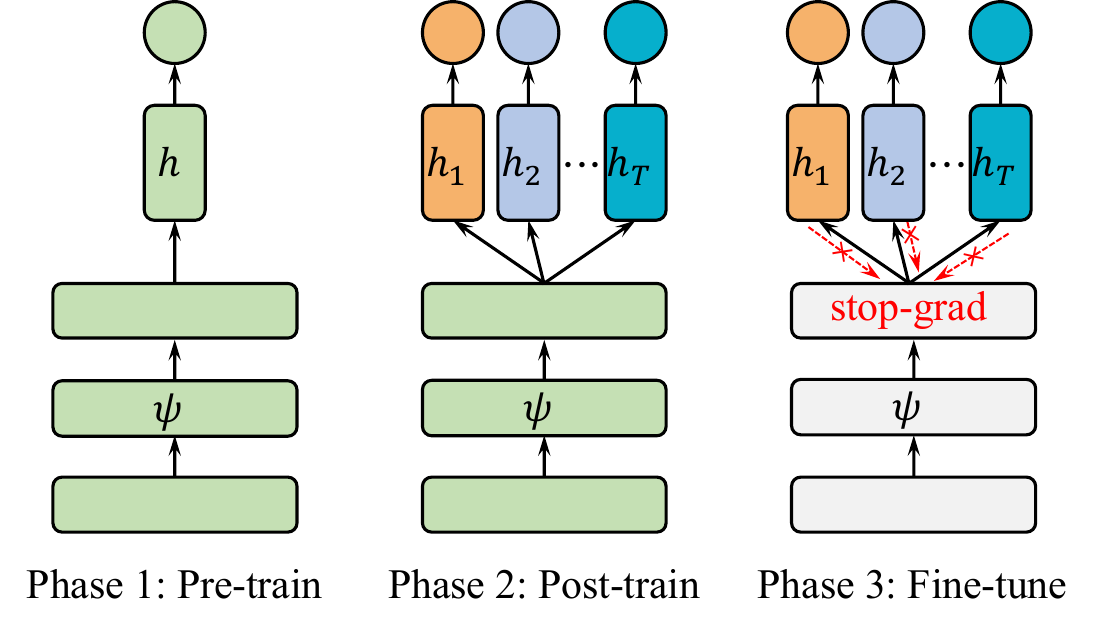}
	}
	\subfigure[Training curves of each phase]{
		\label{train_process}
		\includegraphics[width=0.3465\textwidth]{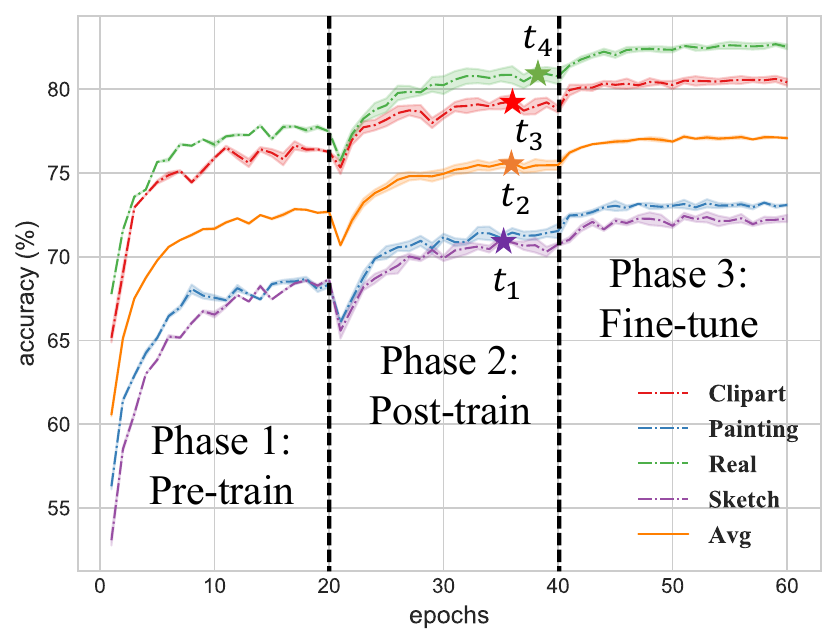}
	}
	\caption{{(a)}: Explanations of different phases of D-Train. $\psi$ denotes the feature extractor; $h$ denotes the shared head in the pre-training phase; $\{{h}_{1}, {h}_{2},..., {h}_{T}\}$ denote the domain-specific heads at the next two phases. During the fine-tuning phase, the parameters of the feature extractor are fixed.  {(b)}: The training curves of each phase of D-Train on various domains: {Clipart}, {Painting}, {Real} and {Sketch} as shown in dotted lines, while the average accuracy over domains is shown in solid line.}

	\label{fig:each-phase}
	
\end{figure*}

\subsection{Pre-train: Warm Up a Root Model}
\label{pre_train}

The power of deep learning models is unleashed by large-scale datasets. However, as mentioned in Section~\ref{intro}, the distributions of sample numbers across domains are usually twisted, imbalanced, or even long-tailed in real-world applications. In this case, some tailed domains may only have limited samples and it would be difficult to train satisfied models on them. To alleviate this problem,  D-Train first pre-trains a single model on samples from all domains to warm up a root model for all domains, especially for the tailed domain with limited samples. The optimization function of the pre-train phase in multiple domains can be formalized as
\begin{equation}
	{\psi}_0, {h}_0 =  \mathop{\arg\min }\limits_{\psi, h} {\frac{1}{T} \sum_{t=1}^T }  \frac{1}{n_t}  \sum_{i=1}^{n_t} \mathcal{L}\left[(h \circ \psi)(\mathbf{x}_i^t), \mathbf{y}_i^t \right],
\end{equation}
where $h$ and $\psi$ are the \emph{shared} head and backbone at the pre-training phase. After the training process converges, $\psi_0$ and ${h}_0$ will be good initializations for the next phase, as shown in Figure~\ref{phase}. To verify it, we take DomainNet as an example and show the training curves of the proposed method in Figure~\ref{train_process}. Note that, the experiments are repeated $5$ times to show both mean and standard deviation.

\subsection{Post-Train: Split Into Multi-Heads}
\label{post_train}

As mentioned before,  jointly training a single model obscures domain distinctions, leading to domain conflict caused by the specificity of different domains. To reflect the domain specificity and achieve satisfactory performance for each domain, we adopt the \emph{shared-bottom} architecture that has a shared feature extractor $\psi$ and various domain-specific heads $\{{h}_{1}, {h}_{2},..., {h}_{T}\}$ to tackle the challenge of \textit{dataset bias} across domains. With this design, the parameters of the feature extractor will be updated simultaneously by gradients of samples from all domains, but the parameters of domain-specific heads are trained on each domain.
The optimization function of the post-training phase is formalized as
\begin{equation}
	\widetilde{\psi}, \{\widetilde{h}_{1}, \widetilde{h}_{2},..., \widetilde{h}_{T}\}   =  \mathop{\arg\min }\limits_{\psi, \{{h}_{1}, {h}_{2},..., {h}_{T}\}} {\frac{1}{T} \sum_{t=1}^T }  L_t 
\end{equation}
\begin{equation}
	L_t  = \frac{1}{n_t}  \sum_{i=1}^{n_t} \mathcal{L}\left[(h_t \circ \psi)(\mathbf{x}_i^t), \mathbf{y}_i^t \right]
\end{equation}
where the domain-agnostic feature extractor $\psi$ is initialized as ${\psi}_0$ while each domain-specific head $h_t$ of domain $t$ is initialized as ${h}_0$. 
After the training process converges, $\psi$ and ${h}$ will reach strong points of $\widetilde{\psi}$ and $\widetilde{h}$ as shown in Figure~\ref{train_process}. Since the shared-bottom architecture contains both domain-agnostic and domain-specific parameters, the training across domains maintains a lukewarm relationship. In this way, the challenge of \textit{dataset bias} across domains and \textit{domain domination} can be somewhat alleviated. The training curves as shown in Figure~\ref{train_process} also witness a sharp improvement via splitting into multi-heads across domains. Note that, at the beginning of the post-train phase, the test accuracy of each domain drops first owing to the training mode switches from fitting all domains to each specific domain.

\subsection{Fine-tune: Decouple-Train for Independence}
\label{fine_tune}
\begin{figure}[!htp]
	\centering
	\subfigure[After Post-train]{
		\label{weak_decoupling}
		\includegraphics[width=0.22\textwidth]{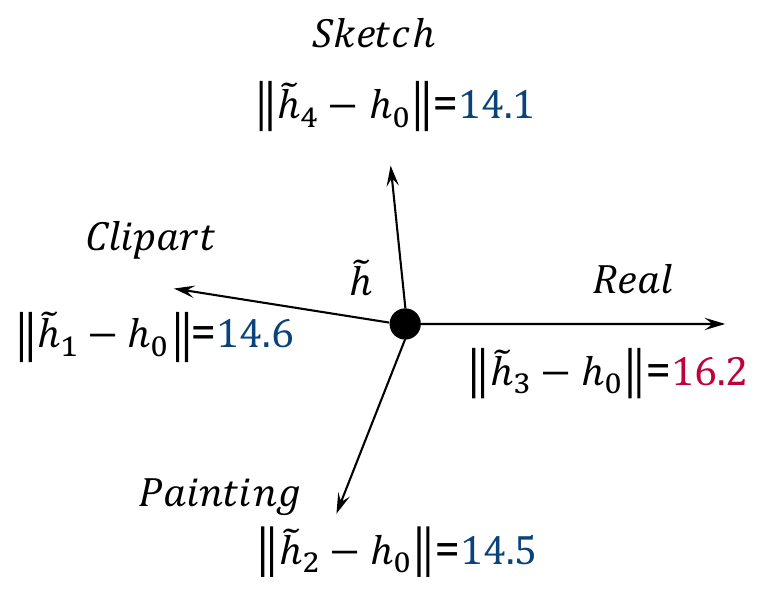}
	}
	\subfigure[After Fine-tune]{
		\label{strong_decoupling}
		\includegraphics[width=0.22\textwidth]{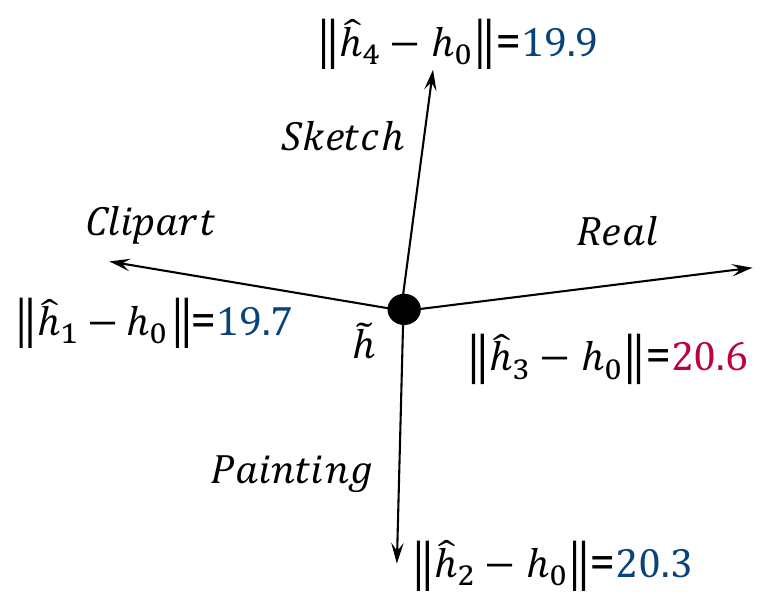}
	}
	\caption{Visualization on domain domination where ${h}_0$ is the domain-agnostic head at the pre-training phase. $\| \cdot \|$ denotes the Euclidean norm of the parameter update.}
	\label{fig:domain_domination}
\end{figure}

Regarding the benefits of domain-specific parameters across domains, a natural question arises: Can the shared-bottom architecture fully solve the problem of domain domination? To answer this question, we adopt the Euclidean norm to calculate the parameter update between the domain-specific heads $\{{h}_{1}, {h}_{2},..., {h}_{T}\}$ and the domain-agnostic head ${h}_0$ at the pre-training phase on DomainNet. As revealed in Table~\ref{table:DomainNet}, domain \texttt{Real} has more examples than other domains and is believed to dominate the training process. Our analysis in Figure~\ref{weak_decoupling} also verified that domain \texttt{Real} ($h_3$) is far away from the initial head. Since the parameter update at each training step $i$ is $h_t^{(i+1)} = h_t^{(i)} - \eta \frac{\partial L}{\partial h_t^{(i)}}$, we can easily attain the {cumulative} parameter update when the training converges: $h_t^{(T)} = h_t^{(0)} - \eta \sum_{i=0}^{T-1} \frac{\partial L}{\partial h_t^{(i)}}$, by repeating the calculation as $h_t^{(1)} = h_t^{(0)} - \eta \frac{\partial L}{\partial h_t^{(0)}}, h_t^{(2)} = h_t^{(1)} - \eta \frac{\partial L}{\partial h_t^{(1)}}, \cdots,  h_t^{(T)} = h_t^{(T-1)} - \eta \frac{\partial L}{\partial h_t^{(T-1)}}$. Here, domain \texttt{Real} ($h_3$) has the largest value of the {{cumulative}} parameter update, we can say that it still \textit{dominates} the training process. Hence, though the challenge of dataset bias across domains can be alleviated by introducing domain-specific heads in the shared-bottom architecture, the domain domination problem still exists after the post-train phase. This is reasonable since the domain-shared parameters of the backbone will be dominated by the head domains. To this end, we propose a decoupling training strategy by fully fixing the parameters of the feature extractor to achieve domain independence. Formally, 
\begin{equation}
	\widehat{h}_t= \mathop{\arg\min }\limits_{h_t} \frac{1}{n_t}  \sum_{i=1}^{n_t}\mathcal{L}\big[(h_t \circ \widetilde{\psi})(\mathbf{x}_i^t), \mathbf{y}_i^t\big], \quad t = 1,2,...,T.
\end{equation}
In this way, the parameters of the domain-specific heads will be learned by samples from each domain. With this kind of \textit{domain-independent} training, the head domains will no longer dominate the training of the tailed domains at this phase. Further, the parameter update between phases becomes more balanced across domains as shown in Figure~\ref{strong_decoupling}. Meanwhile, the training curves as shown in Figure~\ref{train_process} reveal that fine-tuning across domains further improves performance.

\subsection{Why Does D-Train Work?}

As mentioned before, dataset bias and domain domination are the main challenges of multi-domain learning. All of the existing works, including the methods from the perspectives of both {seeking commonalities} or {reserving differences}, have to face the challenge of \textit{seesaw effect}. With the shared parameters across domains, these methods will influence each other. Specifically, as shown in Figure~\ref{train_process},{different domains achieve the optimal performance at different time stamps}.
When the problem of domain domination or domain conflict caused by dataset bias cannot be ignored, the MDL model will struggle to find an optimal solution for all domains. On the contrary, with the proposed decoupling training strategy, different domains in D-Train will train independently at the fine-tuning phase, enabling it the flexibility to
achieve \textit{an optimal solution for each domain at different time stamps}.

\begin{figure}[tp]
	\subfigure[Shared Bottom]{
		\includegraphics[width=0.46\textwidth]{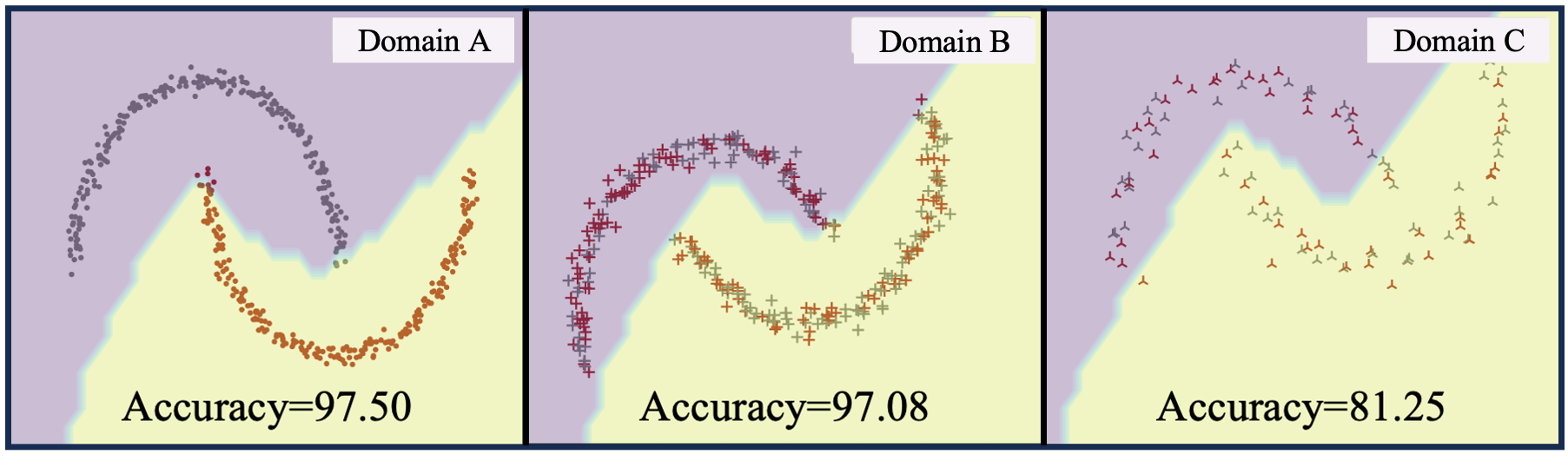}
	}
	\subfigure[D-Train]{
		\includegraphics[width=0.46\textwidth]{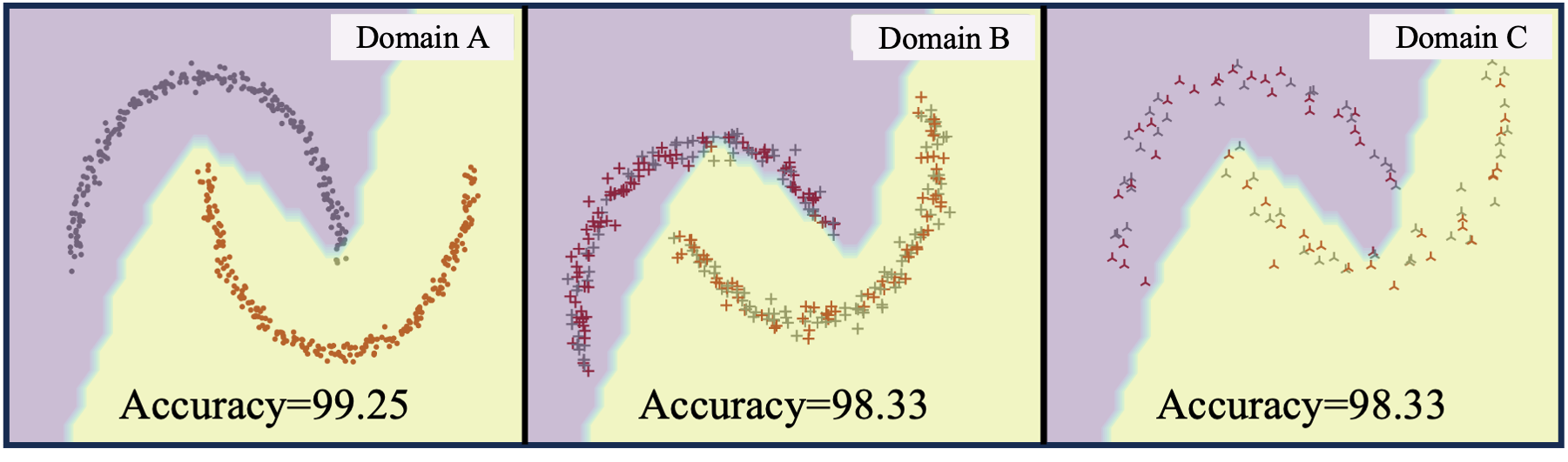}
	}
	\caption{Decision boundaries on Two-Moon, where numbers in the legend indicate the accuracy of each domain.}
	\label{two_moon}
\end{figure}

Further, we visualized the decision boundaries of different domains on \textit{Two-Moon} with different scales and distributions in Figure~\ref{two_moon}. Caused by dataset bias and domain domination, the decision boundary of existing methods like Shared Bottom have some \textit{conflict areas}, while D-Train can further adjust them by domain-independent fine-tuning.

\begin{table*}[!h]
	\addtolength{\tabcolsep}{1.5pt}
	\centering
	% \vspace{5pt}
	\begin{tabulary}{\linewidth}{l|*{4}c|cc}
		\toprule
		Method & Art & Clipart & Product & Real-World & Avg. Acc. & Worst Acc.  \\
		\midrule
		\#Samples & 2427 & 4365 & 4439 & 4357 & - & - \\
		\midrule
		Separatly Train & 76.8 & 77.2 & 92.9 & 87.1 & 83.5 & 76.8 \\
		Jointly Train & 73.5 & 73.3 &  91.4 & 86.7 & 81.2 & 73.3\\ 
		MulANN~\cite{MulANN} 
		& 77.8  & 80.0&	92.5 & 87.3 & 84.4 & 77.8 \\
		DANN-MDL~\cite{ganin2016domain} &  75.9 & 77.2 & 92.2 & 87.2 & 83.1 & 75.9\\
		ASP-MDL~\cite{liu2017adversarial} 
		& 78.8 & 79.3 & 93.2 & 87.5 & 84.7 & 78.8 \\
		CDAN-MDL~\cite{Long18CDAN} & 78.6 & 79.0 & 93.6 & 89.1 & 85.1 & 78.6 \\ 
		Shared Bottom~\cite{ruder2017overview} & 76.5 &	{80.2} & 93.8 &88.8 & 84.8 & 76.5 \\
		MoE~\cite{jacobs1991adaptive} & 76.8 & 77.1 & 92.3 & 87.4 & 83.4 & 76.8  \\
		MMoE~\cite{ma2018modeling} & 78.8 & 79.6 & 93.4 & 88.9 & 85.2 & 78.8 \\
		PLE~\cite{tang2020progressive}& 78.0 & 79.8 & 93.6 & 88.5 & 85.0& 78.0  \\
		\textbf{D-Train (ours)} & \textbf{80.0} & \textbf{80.3}  & \textbf{94.1}  & \textbf{89.5} & \textbf{86.0} & \textbf{80.0}\\ 
		\bottomrule
	\end{tabulary}
	\caption{Accuracy (\%) on {{Office-Home} }for multi-domain learning (ResNet-50).}
	\label{table:office_home}
\end{table*}
\section{Experiments}	
In this section, we compared the proposed D-Train method with baselines. Note that, domains are {balanced} by setting a fixed `\textit{batch\_size}' for every domain, \textit{i.e.}, the `\textit{batch\_size}' for each domain is $20$ if there are $80$ examples for $4$ domains in a mini-batch. We adopted this \textit{balanced} version for all baselines and the proposed method because we found that the balanced implementation works much better than the original imbalanced one, \textit{e.g.}, Shared Bottom achieves an average accuracy of $75.9$ over $74.1$ in two versions.
\begin{table*}[!h]
	\addtolength{\tabcolsep}{2.5pt}
	\centering
	% \vspace{5pt}
	\begin{tabulary}{\linewidth}{l|*{4}c|cc}
		\toprule
		Method & Clipart & Painting & Real & Sketch & Avg. Acc. & Worst Acc. \\
		\midrule
		\#Samples & 49k & 76k & 175k & 70k & - & - \\
		
		% \#Samples & 48837 & 75759 & 175327 & 70386 & - & - \\
		
		\midrule
		Separatly Train & 78.2 & 71.6 & \textbf{83.8} & 70.6 & 76.1 & 70.6 \\
		Jointly Train & 77.4 & 68.0 & 77.9 & 68.5 & 73.0 & 68.0 \\ 
		MulANN~\cite{MulANN} 
		& 79.5 & 71.7 &	81.7 & 	69.9 & 75.7 & 69.9  \\
		DANN-MDL~\cite{ganin2016domain} & 79.8 &71.4 &81.4 &	70.3 & 75.7 & 70.3 \\
		ASP-MDL~\cite{liu2017adversarial} 
		& 80.1 & 72.1 & 81.2 &	70.9 & 76.1 & 70.9 \\
		CDAN-MDL~\cite{Long18CDAN} & 80.2 &72.2 &	81.3 & 71.0 & 76.2 & 71.0   \\ 
		Shared Bottom~\cite{ruder2017overview} & 79.9 & 72.1 & 81.9 & 69.7 &	75.9 & 69.7   \\
		MoE~\cite{jacobs1991adaptive} & 79.1 	& 70.2 & 79.8  & 69.4 & 74.6  & 69.4 \\
		MMoE~\cite{ma2018modeling} & 79.6 & 72.2 & 82.0	 & 69.8 & 75.9 & 69.8 \\
		PLE~\cite{tang2020progressive}& 80.0 & 72.2 & 82.1 & 70.0 & 76.1   & 70.0\\
		\midrule
		D-Train (w/o Fine-tune) & 79.9 & 71.3 & 81.3 & 70.9 & 75.9 & 70.9 \\ 
		D-Train (w/o Post-train)&
		79.8 & 72.9 & 	81.7 & 72.1 & 76.6  & 72.1 \\ 
		D-Train (w/o Pre-train) & 80.9 & 72.9 & 82.7 & 71.5 & 77.0  & 71.5\\
		\textbf{D-Train (ours)} & \textbf{81.5} & \textbf{72.8} & 82.7 & \textbf{72.2} & \textbf{77.3} & \textbf{72.2} \\ 
		\bottomrule
	\end{tabulary}
	\caption{Accuracy (\%) on {{DomainNet}} for multi-domain learning (ResNet-101).}
	\label{table:DomainNet}
\end{table*}

\subsection{Standard Benckmarks}

In this section, we adopt two standard benchmarks in the field of domain adaptation with various dataset scales, where Office-Home is in a low-data regime and DomainNet is a large-scale one. 

% D-Train and all baselines in this section are implemented in a popular open-sourced library~\footnote{https://github.com/thuml/Transfer-Learning-Library} which has implemented a high-quality code base for many domain adaptation baselines.

%	\vspace{-5pt}
\paragraph{Low-data Regime: Office-Home} Office-Home is a standard multi-domain learning dataset ~\cite{Venkateswara17Officehome} with $65$ classes and $15,500$ images from four significantly different domains: Art, Clipart, Product, and Real-World. As shown in Figure~\ref{office_home_samples}, there exist challenges of dataset bias and domain domination in this dataset. Following existing works on this dataset, we adopt ResNet-50 as the backbone and randomly initialize fully connected layers as heads. We set the learning rate as $0.0003$ and batch size as $24$ in each domain for  D-Train and all baselines.

%	The distribution of sample numbers across domains is twisted and the domain of "Art" has only about a half of the samples as other domains.

As shown in Table~\ref{table:office_home}, Separately Training is a strong baseline and even outperforms Jointly Training, since the latter obscures domain distinctions and cannot tackle the dataset bias across domains. Multi-domain learning methods from the perspective of domain alignment work much better by introducing a domain discriminator and exploiting the domain information. However, applying domain alignment is not an optimal solution in MDL since the domain gap can only be reduced but not removed. Finally, the proposed D-Train consistently improves on all domains, even the tailed domain of ``Art''.  D-Train achieves a new state-of-the-art result with an average accuracy of $86.0$.

%	Surprisingly, most of the methods that adopt multiple branch architecture achieve fair performance

\paragraph{Large-Scale Dataset: DomainNet} DomainNet~\cite{peng2018moment} is a large-scale multi-domain learning and domain adaptation dataset with $345$ categories. We utilize $4$ domains with different appearances including \textit{Clipart}, \textit{Painting}, \textit{Real}, and \textit{Sketch} where each domain has about $40,000 $ to $200,000$ images. Following the code base in the Transfer Learning Library, we adopt mini-batch SGD with the momentum of $0.9$ as an optimizer, and the initial learning rate is set as $0.01$ with an annealing strategy. We adopt ResNet-101 as the backbone since DomainNet is much larger and more difficult than the previous Office-Home dataset. Meanwhile, the batch size is set as $20$ in each domain here for D-Train and all baselines.

As shown in Table~\ref{table:DomainNet}, the proposed D-Train outperforms all baselines, no matter measured by average accuracy or worst accuracy in all domains.  Since the domain of ``Real'' has much more samples than other domains, it achieves competitive performance while separately training. However, other domains benefit from training via D-Train.	

% ~\footnote{https://github.com/p-lambda/wilds} 
\subsection{Applications of Satellite Imagery}	
In this section, we adopt a popular satellite dataset named Functional Map of the World (FMoW) ~\cite{christie2018functional}, which aims to predict the functional purpose of buildings and land use on this planet, contains large-scale satellite images with different appearances and styles from various regions:  \textit{Africa}, \textit{Americas}, \textit{Asia}, \textit{Europe}, and \textit{Oceania}. FMoW is a natural dataset for multi-domain learning. D-Train and all baselines in this section are implemented in a popular open-sourced library named WILDS since it enables easy manipulation of this dataset. Each input $\mathbf{x}$ in FMoW is an RGB satellite image that is resized to $224 \times 224$ pixels and the label $y$ is one of $62$ building or land use categories. 
For all experiments, we follow \cite{christie2018functional} and use a DenseNet-121 model \cite{Huang_2017_CVPR} pre-trained on ImageNet. We set the batch size to be $64$ on all domains.
Following WILDS, we report the average accuracy and worst-region accuracy in all multi-domain learning methods.

\begin{table*}[!h]
	\addtolength{\tabcolsep}{-0.5pt}
	\centering
	% \vspace{5pt}
	\begin{tabulary}{\linewidth}{l|*{5}c|cc}
		\toprule
		Method & Asia & Europe & Africa & America & Oceania & Avg. Acc. & Worst Acc. \\
		\midrule
		\#Samples & 115k & 182k & 37k & 176k & 15k & - & -\\
		% #Samples & 115039 & 181602 & 36723 & 175598 & 14617 & - & -\\
		\midrule
		Separately Train & 59.4 & 56.9 & 72.9 & 59.4 & 65.1 & 58.7 & 56.9 \\
		Jointly Train & 60.8 & 57.2 & \textbf{77.0} & \textbf{63.5} & 71.6 & 60.4 & 57.2 \\
		MulANN~\cite{MulANN} & 61.2 & 57.5 & 74.6 & 62.3 & 67.9 & 60.2 & 57.5 \\
		DANN-MDL~\cite{ganin2016domain} & 55.9 & 55.5 & 61.9 & 58.2 & \textbf{74.2} & 56.7 & 55.5 \\
		ASP-MDL~\cite{liu2017adversarial} 
		&  54.5 & 53.9 & 73.4 & 57.4 & 70.3 & 55.3 & 53.9 \\
		CDAN-MDL~\cite{Long18CDAN} & 57.0 & 56.8 & 68.0 & 59.7 & 70.3 & 57.7 & 56.8 \\
		Shared Bottom~\cite{ruder2017overview} & 58.1 & 57.1 & 75.4 & 61.3 & 71.9 & 59.8 & 57.1 \\
		MoE~\cite{jacobs1991adaptive} &  55.9 & 54.0 & 63.2 & 59.0 & 70.6 & 56.3 & 54.0 \\
		MMoE~\cite{ma2018modeling} & 60.7 & 55.7 & 65.6 & 62.4 & 64.8 & 58.7 & 55.7 \\
		PLE~\cite{tang2020progressive}&  58.2 & 56.5 & 74.6 & 61.7 & 72.5 & 59.0 & 56.5 \\
		\textbf{D-Train (ours)} & \textbf{62.3} & \textbf{58.3} & \textbf{77.0} & 62.7 & 68.3 & \textbf{61.0} & \textbf{58.3} \\
		\bottomrule
	\end{tabulary}
	\caption{Accuracy (\%) on {{FMoW}} for multi-domain learning (DenseNet-121).}
	\label{table:FMoW}
\end{table*}

As shown in Table~\ref{table:FMoW}, it’s not wise to train a separate model for each domain on FMoW, since the data 
on some domains is extremely scarce. FMoW is a difficult and scarce dataset, \textit{e.g.}, Oceania has only 15k images, causing some categories to involve only a dozen examples.
Joint Train is also not optimal, because conflicts widely exist in some domains. Note that D-Train also outperforms all baselines in this difficult dataset. 
% However, {the gap across domains is not as large as other datasets}, making the tail domain benefit more from aligning feature distributions by DANN-MDL.

\begin{table*}[tb]
	% \addtolength{\tabcolsep}{-2.5pt}
	\centering
	% \small
	\begin{tabular}{l|ccccccc|cc}
		\toprule
		
		Method                   & Books           & Elec.    & {TV}   & {CD}   & {Kindle}    & {Phone} & {Music}   & {$\texttt{AUC}_d$} & {$\texttt{AUC}_s$}\\ 
		
		\midrule
		\#Samples                & 19.2M          & 3.70M           & 3.28M           & 2.96M           & 2.25M           & 0.81M                       & 0.38M           &     --  &      --             \\ \midrule
		Separately Training      & 66.09           & 77.50           & 79.43           & 59.69           & 52.79           & 70.06                       & 52.95           & 65.50              & 67.17             \\
		Jointly Training           & 69.01           & 78.87           & 85.06           & 64.24           & 59.15           & 69.89                       & 49.71           & 67.99              & 70.43             \\
		MulANN~\cite{MulANN}       & 68.95           & 78.90           & 84.56           & 64.79           & 58.64           & 70.43                       & 52.13           & 68.34              & 70.40             \\
		DANN-MDL~\cite{ganin2016domain}      & 68.64           & 80.33           & 86.08           & 66.32           & 58.59           & 72.47                       & 54.21           & 69.52              & 70.74             \\
		CDAN-MDL~\cite{Long18CDAN}       & 69.74           & 80.63           & 85.88           & 67.24           & 60.61           & 73.34                       & 57.39           & 70.69              & 71.69             \\
		MoE~\cite{jacobs1991adaptive}                       & 73.51           & 85.88           & 89.66           & 74.94           & 63.45           & 79.63                       & 66.08           & 76.16              & 76.04             \\ \midrule
		Shared Bottom~\cite{ruder2017overview}          & 70.91           & 74.87           & 85.51           & 67.18           & 60.56           & 74.59                       & 59.14           & 70.39              & 71.73             \\
		Shared Bottom + D-Train      & 71.35           & 74.76           & 85.52           & 67.61           & 60.20            & 73.53                       & 61.61           & 70.65              & 71.99             \\
		% Shared Bottom + AT (ours) & 70.33           & 80.42           & 87.13           & 71.97           & 61.34           & 80.09                       & 61.48           & 73.25              & 72.83             \\
		\midrule
		MMoE~\cite{ma2018modeling}                     & 73.67           & 86.15           & 89.23           & 75.50           & 62.43           & 81.91                       & 63.69           & 76.01              & 76.13             \\
		MMoE + D-Train          & 74.50           & 86.09           & 88.60           & 77.13           & 66.46           & 82.13                       & 69.01           & 77.70              & 77.05             \\
		% MMoE + AT (ours)       & 74.60           & 86.89           & 87.64           & 77.30           & 67.09           & 82.80                       & \textbf{70.06}  & 78.05              & 77.18             \\
		\midrule
		PLE~\cite{tang2020progressive}                     & {75.25}  & 85.36           & 88.54           & 76.09           & {69.35}  & 81.02                       & 67.75           & 77.62              & 77.46             \\
		PLE  + D-Train              & 74.70           & 86.70           & 89.53           & {77.40}  & 69.26           & 82.47                       & 69.91           & 78.57              & 77.56             \\
		% PLE + AT (ours)       & 75.03           & \textbf{87.38}  & \textbf{90.28}  & 77.11           & 69.17           & \textbf{83.32}              & 69.93           & \textbf{78.89}     & \textbf{77.90}    \\ 
		\bottomrule
	\end{tabular}
	\normalsize
	\caption{AUC (\%) on Amazon Product Review for multi-domain learning.}
	\label{amazon_results}
\end{table*}

% ~\footnote{https://jmcauley.ucsd.edu/data/amazon/} 
\subsection{Applications of Recommender System}

We adopt a popular dataset named Amazon Product Review (\textit{Amazon}) which is a large crawl of product reviews from the \textit{Amazon} website, recording users' \textit{preferences} (by score) for different products. We select $7$ typical subsets with various scales including Books (\textit{Books}), Electronics (\textit{Elec.}), Movies\_and\_TV (\textit{TV}), CDs\_and\_Vinyl (\textit{CD}), Kindle\_Store (\textit{Kindle}), Cell\_Phones\_and\_Accessories (\textit{Phone}), Digital\_Music (\textit{Music}). As shown in Tabel~\ref{amazon_results}, different domains have various samples from $0.38M$ to $19.2M$. For each domain,
diverse user behaviors are available, including more than $5$ reviews for each user-goods pair. The features consist of \texttt{goods\_id} and \texttt{user\_id}. Users in these domains have different preferences for various goods.

% D-Train and all baselines in this section are implemented based on a popular open-sourced library named pytorch-fm~\footnote{https://github.com/rixwew/pytorch-fm}.
We use DNN as the CTR method and the \texttt{embed\_dim} and \texttt{mlp\_dim} are both set as $16$. The layer number of the expert and the tower are set as $2$ and $3$ respectively.
We report AUC (Area Under the Curve) for each domain. Further, \texttt{AUC\_d}  and \texttt{AUC\_s} are averaged over all domains and all samples respectively to intuitively compare D-Train with other baselines. For all models, we use Adam as the optimizer with exponential decay, in which the learning rate starts at $1e^{-3}$ with a decay rate of $1e^{-6}$. During training, the mini-batch size is set to $2048$.
As shown in Table~\ref{amazon_results}, D-Train
yields larger improvements than a variety of MDL baselines.
% on Amazon Product Review dataset.

\subsection{Plug-in Unit}
D-Train can be used as a general plug-in unit for existing
MDL methods. When D-Train is used as a plugin unit, MMoE and PLE \textit{add an extra phase} to only train the parameters of domain-specific heads while fixing the other parameters. As shown in Table~\ref{amazon_results}, D-Train
can further improve these competitive MDL methods on Amazon Product Review dataset, by tailoring D-Train into them.

\subsection{Ablation Study}
As shown in Table~\ref{table:DomainNet}, ablation study on DomainNet by removing each phase respectively reveals that only utilizing all of these phases works best. In particular, D-Train (w/o Fine-tune) works much worse than other ablation experiments, which reveals the importance of decoupling training for domain independence.

\subsection{Online Development}

We empirically evaluate D-Train in the online advertising scenario of Tencent's DSP, which needs to serve billions of requests per day. MMoE-like architecture is utilized to predict pCVR for various conversion tasks, such as purchase and follow. Hundreds of features are used in total, including user behavior features, ad side features, and context side features. Online A/B testing from early July 2022 to August 2022 demonstrates that D-Train achieves $0.36\%$ cost lift and $1.69\%$ GMV(normal) lift over the online baseline, where the improvements on different domains are $3.71\%$, $-0.08\%$ (not statistically significant) and $2.03\%$, respectively.
% We estimate these performance gain leads to hundreds of millions of dollars in revenue lift per year.
% By developing D-Train, we aim to improve the performance of various scenarios of DSP
% Numerous efforts have been made for it but most of the works focus on a special application and a general approach is missing.
\section{Conclusion}	
Multi-domain learning (MDL) is of great importance in both academia and industry. In this paper, we explicitly uncover the main challenges of dataset bias and domain domination in multi-domain learning, especially the latter since it is usually ignored in most existing works. We further propose a frustratingly easy and hyperparameter-free multi-domain learning method named Decoupled Train (D-Train) that highlights the domain-independent fine-tuning to alleviate the obstacles of \textit{seesaw effect} across multiple domains. 
% Despite its extraordinary simplicity and efficiency, D-Train performs remarkably well in extensive evaluations of various datasets from standard benchmarks to applications of satellite imagery and recommender system.

\section{Acknowledgements}
We would like to thank many colleagues, in particular, Junguang Jiang, Yiwen Qiu, and Prof. Mingsheng Long for their valuable discussions and efforts. Further, we would like to thank Rong Chen, Yuxuan Han, and Wei Xue for their successful deployments of this work into our online system, with a significant improvement in the online A/B test.

\bibliography{main}

\clearpage
\appendix

\section{A. Experiment Details}
\subsection{A.1. Baselines}  

\begin{figure*}[tp]
	\centering
	\subfigure[Joint]{
		\label{joint}
		\includegraphics[width=0.117\textwidth]{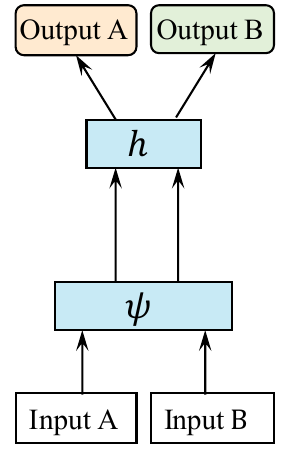}
	}
	\subfigure[MulANN]{
		\label{MulANN}
		\includegraphics[width=0.146\textwidth]{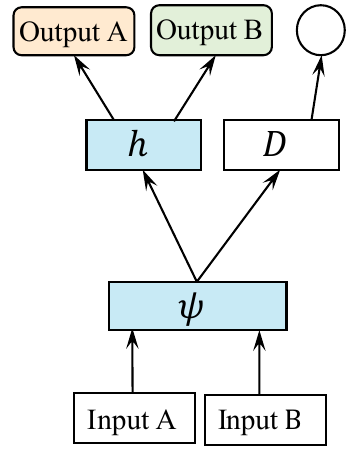}
	}
	\subfigure[DANN-MDL]{
		\label{DANN-MDL}
		\includegraphics[width=0.173\textwidth]{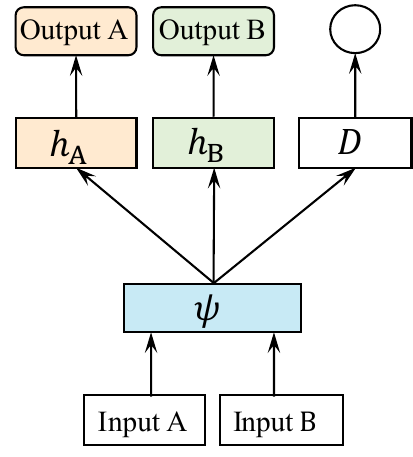}
	}
	\subfigure[ASP-MTL]{
		\label{ASP-MTL}
		\includegraphics[width=0.226\textwidth]{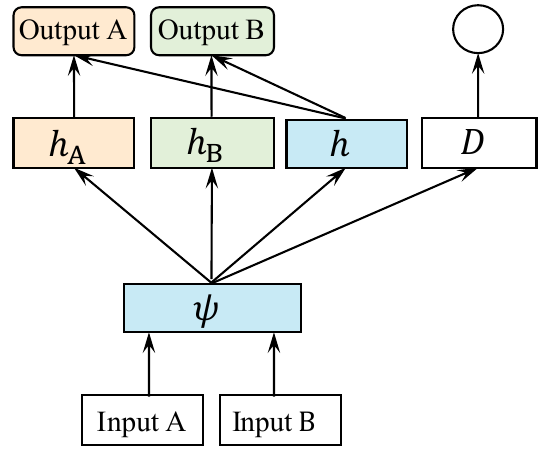}
	}
	\subfigure[CDAN-MDL]{
		\label{CDAN-MDL}
		\includegraphics[width=0.226\textwidth]{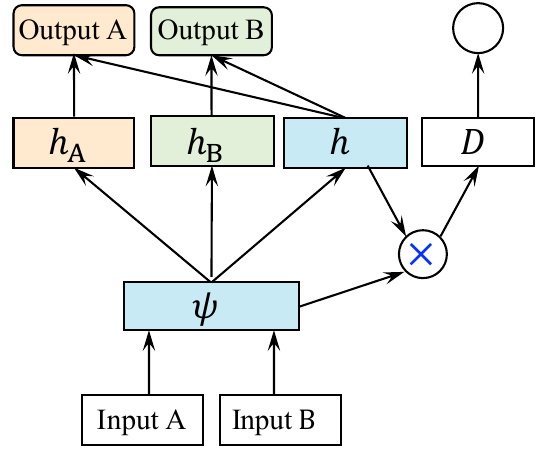}
	}
	\subfigure[Separate]{
		\label{separate}
		\includegraphics[width=0.16\textwidth]{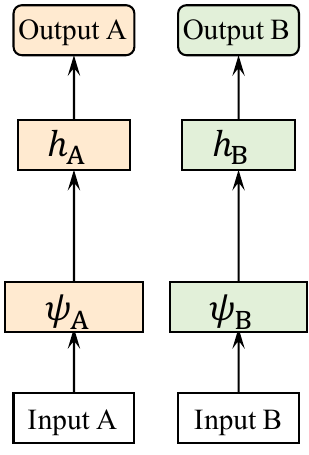}
	}
	\subfigure[SB]{
		\label{SB}	\includegraphics[width=0.144\textwidth]{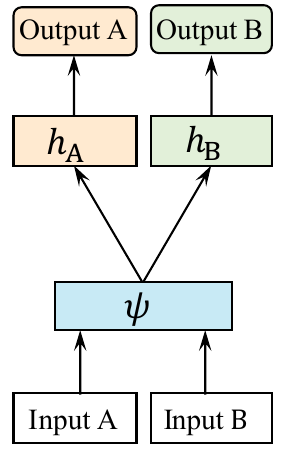}
	}
	\subfigure[MoE]{
		\label{MoE}
		\includegraphics[width=0.18\textwidth]{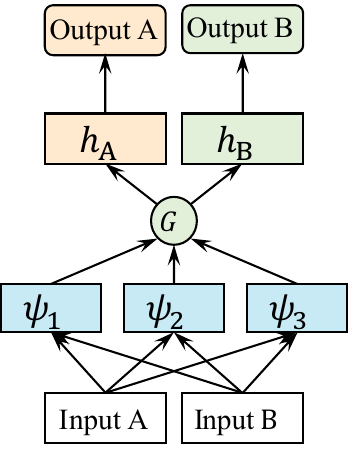}
	}
	\subfigure[MMoE]{
		\label{MMoE}
		\includegraphics[width=0.18\textwidth]{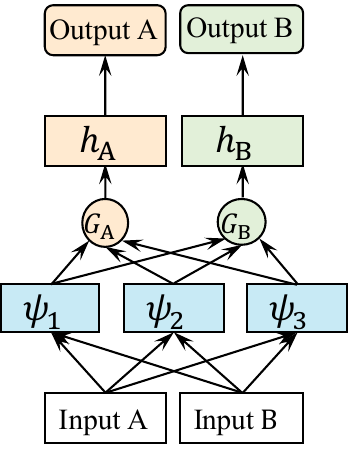}
	}
	\subfigure[PLE]{
		\label{PLE}
		\includegraphics[width=0.18\textwidth]{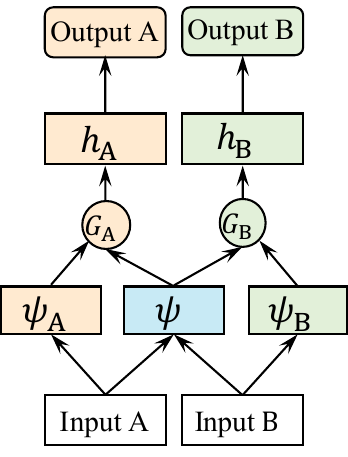}
	}
	\caption{Network architectures of various multi-domain learning methods in which methods at the top row tackle dataset shift from the perspective of domain alignment and methods at the bottom row adopt multi-branch architecture with domain-specific and domain-shared parameters. The used terminologies include feature extractor (backbone) $\psi$, classifier (head) $h$, domain discriminator $D$, and gate $G$. The blue items represent domain-shared modules while others are domain-specific modules.}
	\label{fig:baseline_comparison}
\end{figure*}

We use {PyTorch}\footnote{\url{http://pytorch.org}} as the deep learning framework and run all methods in Titan V. We compared the proposed D-Train method with three categories of baselines: Single Branch, Domain Alignment, and Multiple Branch. The implementation details of these baselines are summarized as follows: 
\begin{itemize}
	\item\textbf{Separately Train} is a strong baseline that separately trains on each domain when enough data is available as shown in Figure~\ref{separate}. Here, each domain is trained independently and samples from other domains will not influence the target domain. However, separately training multiple models by domains reduces training data in each model~\cite{when_do_domain_matters}.
	\item \textbf{Jointly Train} trains a single model with samples from all domains regardless of the domain information as shown in Figure~\ref{joint}.  In this case, jointly training a single model obscures domain distinctions~\cite{when_do_domain_matters} and it may bring the challenge of dataset bias.
	\item \textbf{DANN-MDL}~\cite{ganin2016domain} is a direct extension of DANN in domain adaptation into multi-domain learning as shown in Figure~\ref{DANN-MDL}. Here, we introduce a domain discriminator for the original MDL model to enable distribution alignment across domains.
	\item \textbf{MulANN}~\cite{MulANN} introduces a domain discriminator into the shared model with a single head, as shown in Figure~\ref{MulANN}. Here, we remove the extra loss in the original paper on unlabeled data since all samples are labeled in our setting.
	\item  \textbf{ASP-MTL}~\cite{liu2017adversarial} includes a domain-agnostic backbone, a domain-shared head, multiple domain-specific heads, and a domain discriminator, as shown in Figure~\ref{ASP-MTL}. Similarly, the loss function specifically designed for natural language processing is removed. 
	\item  \textbf{CDAN-MDL} \cite{Long18CDAN} is a natural extension of CDAN to MDL as shown in Figure~\ref{CDAN-MDL}. Since we empirically found that ASP-MTL is much better than the other two variants of domain alignment methods, we also designed a share-private architecture for CDAN-MDL.
	\item \textbf{Shared Bottom (SB)}~\cite{ruder2017overview} is the frustratingly easy but effective MDL  method, as shown in Figure~\ref{SB}. Here, a shared backbone is included to extract representations while multiple domain-specific heads are designed considering the specificity of each domain.
	\item  \textbf{MoE}~\cite{jacobs1991adaptive} utilizes the insightful idea of the mixture of experts that learns different mixtures patterns of experts assembling as shown in Figure~\ref{MoE}. Here, the number of experts is selected from the set of $[3, 4, 5, 6]$ according to the performance on the validation set.
	\item  \textbf{MMoE}~\cite{ma2018modeling} extends MoE to the multi-gate version and explicitly learns to model task relationships from data by designing a task-specific gating network for each task, as shown in Figure~\ref{MMoE}. Similarly, the number of experts is selected from the set of $[3, 4, 5, 6]$ according to the performance on the validation set of each experiment.
	\item \textbf{PLE}~\cite{tang2020progressive} explicitly separates domain-shared and domain-specific experts as shown in Figure~\ref{PLE}. The PLE model in the original paper adopts a progressive routing mechanism to extract and separate deeper semantic knowledge gradually~\cite{tang2020progressive}. 
\end{itemize}

\begin{table*}[!h]
	% \addtolength{\tabcolsep}{15pt}
	\centering
	\small
	\caption{Comparisons among various MDL methods. Note that, $T$ is the domain number.  $n_{\psi}$ and $n_h$ are  parameter number of the backbone and head respectively. Usually, the inequality $n_{\psi} \gg n_h$ holds. }
	\label{table:para}
	\begin{tabular}{lllc}
		\toprule
		Type & Method & \#Parameters & Hyperparameters \\
		\midrule
		\multirow{2}{*}{Single Branch} & Separately Train & $\mathcal{O}(T \cdot n_{\psi} + T \cdot n_h)$  & W/O \\
		& Jointly Train &  $\mathcal{O}(n_{\psi} + n_h)$ & W/O \\
		\midrule
		\multirow{4}{*}{Domain Alignment}  & MulANN & $\mathcal{O}(n_{\psi} + n_h + n_D) $ & loss trade-off: $\lambda$;  domain discriminator arch. \\
		& DANN-MDL& $\mathcal{O}(n_{\psi} + T \cdot n_h + n_D) $ & loss trade-off: $\lambda$;  domain discriminator arch. \\
		& ASP-MTL& $\mathcal{O}(n_{\psi} + (T + 1) \cdot n_h + n_D) $ & loss trade-off: $\lambda$;  domain discriminator arch. \\
		& CDAN-MDL& $\mathcal{O}(n_{\psi} + (T + 1) \cdot n_h + n_D) $ & loss trade-off: $\lambda$;  domain discriminator arch. \\
		\midrule
		\multirow{4}{*}{Multiple Branch}  &  Shared Bottom & $\mathcal{O}(n_{\psi} + T \cdot n_h)$ & W/O \\
		& MoE & $\mathcal{O}(E \cdot n_{\psi} + n_G + T \cdot n_h )$ & expert number (EN): $E$; gate network arch.\\
		& MMoE & $\mathcal{O}(E \cdot n_{\psi} + T \cdot (n_h + n_G))$ & expert number (EN): $E$; gate network arch.\\
		& PLE & $\mathcal{O}((m_s + \sum_i^T m_i )\cdot n_{\psi} + T \cdot n_h)$ &  domain-share EN: $m_s$; domain-specific EN: $m_i$\\
		\midrule
		The Proposed & \textbf{D-Train} & $\mathcal{O}(n_{\psi} + T \cdot n_h)$ & W/O\\
		\bottomrule
		
	\end{tabular}
	
\end{table*}

\subsection{A.2. Datasets} 

We conducted experiments on various datasets from standard benchmarks including Office-Home and DomainNet, to applications of satellite imagery (FMoW) and recommender system (Amazon) with various dataset scales, in which Office-Home is in a low-data regime and DomainNet is a large-scale one. D-Train and all baselines in this section are implemented in a popular open-sourced library~\footnote{https://github.com/thuml/Transfer-Learning-Library} which has implemented a high-quality code base for many domain adaptation baselines.

\begin{itemize}
	\item \textbf{Office-Home}: 
	Office-Home~\cite{Venkateswara17Officehome} is a standard multi-domain learning dataset with $65$ classes and $15,500$ images from four significantly different domains: Art, Clipart, Product, and Real-World.  There exist challenges of dataset bias and domain domination in this dataset. 
	Following existing works on this dataset, we adopt ResNet-50 as the backbone and randomly initialize fully connected layers as heads. We set the learning rate as $0.0003$ and batch size as $24$ in each domain for  D-Train and all baselines.
	\item{\textbf{DomainNet}}: 
	DomainNet~\cite{peng2018moment} is a large-scale multi-domain learning and domain adaptation dataset with $345$ categories. We utilize $4$ domains with different appearances including \textit{Clipart}, \textit{Painting}, \textit{Real}, and \textit{Sketch} where each domain has about $40,000 $ to $200,000$ images. The domain of "\textit{Real}" has much more examples than other domains, and is believed to dominate the training process.
	Following the code base in Transfer Learning Library, we adopt mini-batch SGD with a momentum of $0.9$ as an optimizer, and the initial learning rate is set as $0.01$ with an annealing strategy. We adopt ResNet-101 as the backbone since DomainNet is much larger and more difficult than the previous Office-Home dataset. Meanwhile, the batch size is set as $20$ in each domain here for D-Train and all baselines.
	\item{\textbf{FMoW}}: Functional Map of the World (FMoW) ~\cite{christie2018functional} aims to predict the functional purpose of buildings and land use on this planet.
	It contains large-scale satellite images with different appearances and styles from various regions:  \textit{Africa}, \textit{Americas}, \textit{Asia}, \textit{Europe}, and \textit{Oceania}. FMoW is a natural dataset for multi-domain learning. D-Train and all baselines in this section are implemented in a popular open-sourced library named WILDS~\footnote{https://github.com/p-lambda/wilds} since it enables easy manipulation of this dataset. 
	Each input $\mathbf{x}$ in FMoW is an RGB satellite image that is resized to $224 \times 224$ pixels and the label $y$ is one of $62$ building or land use categories. 
	For all experiments, we follow \cite{christie2018functional} and use a DenseNet-121 model \cite{Huang_2017_CVPR} pretrained on ImageNet. We set the batch size to be $64$ on all domains.
	Following WILDS, we report the average accuracy and worst-region accuracy in all multi-domain learning methods.
	\item{\textbf{Amazon}}:
	In this section, we adopt a popular dataset named Amazon Product Review (\textit{Amazon})~\footnote{https://jmcauley.ucsd.edu/data/amazon/}. We select $7$ typical subsets with various scales including Books (\textit{Books}), Electronics (\textit{Elec.}), Movies\_and\_TV (\textit{TV}), CDs\_and\_Vinyl (\textit{CD}), Kindle\_Store (\textit{Kindle}), Cell\_Phones\_and\_Accessories (\textit{Phone}), Digital\_Music (\textit{Music}). Different domains have various samples from $0.38M$ to $19.2M$. For each domain,
	diverse user behaviors are available, including more than $5$ reviews for each user-goods pair. The features used for experiments consist of \texttt{goods\_id} and \texttt{user\_id}. It is obvious that users in these domains have different preferences for goods.

	D-Train and all baselines in this section are implemented based on a popular open-sourced library named pytorch-fm~\footnote{https://github.com/rixwew/pytorch-fm}. We use DNN as the CTR method and the \texttt{embed\_dim} and \texttt{mlp\_dim} are both set as $16$. The layer number of the expert and the tower are set as $2$ and $3$ respectively.
	We report AUC (Area Under the Curve) for each domain. Further, \texttt{AUC\_d}  and \texttt{AUC\_s} are averaged over all domains and all samples respectively to intuitively compare D-Train with other baselines. For all models, we use Adam as the optimizer with exponential decay, in which the learning rate starts at $1e^{-3}$ with a decay rate of $1e^{-6}$. During training, the mini-batch size is set to $2048$.
\end{itemize}

\section{B. Parameters and Hyperparameters}

Comparisons among various MDL methods are shown in Table~\ref{table:para}. Note that, $T$ is the domain number. $n_{\psi}$ and $n_h$ are  parameter number of the backbone and head respectively. Usually, the inequality $n_{\psi} \gg n_h$ holds. It is obvious that D-Train is \emph{frustratingly easy} and hyperparameter-free, without introducing extra parameters compared with the popular Shared-Bottom method.

\end{document}